\documentclass[11pt]{article}

\usepackage[utf8]{inputenc}
\usepackage{geometry, comment, amsmath, authblk, graphicx, array,caption}
\usepackage{mlmodern, xcolor, soul}
\usepackage[T1]{fontenc}

\usepackage{amsmath,amsfonts}

\newcolumntype{C}[1]{>{\centering\arraybackslash}m{#1}}
\usepackage[hidelinks]{hyperref}
\usepackage[backend=biber,style=apa,sorting=none]
{biblatex} 
\usepackage{setspace, multirow, booktabs, float}
\newcommand{\rdmwidth}{5cm}
\newcommand{\rdmheight}{4.5cm}

\newcommand{\textureimgs}{2.2cm}
\newcommand{\rev}[1]{#1}
\usepackage{lineno}

\newcommand{\myTitle}{Perceptual misalignment of texture representations in convolutional neural networks}

\hypersetup{%
    colorlinks=false, linktocpage=false, pdfborder={0 0 0},%
    pdfstartview=FitV,%
    breaklinks=true, pageanchor=true, pdfpagemode=UseOutlines,%
    plainpages=false, bookmarksnumbered, bookmarksopen=true, bookmarksopenlevel=1,%
    hypertexnames=true, pdfhighlight=/O,
    urlcolor=webbrown, linkcolor=RoyalBlue, citecolor=webgreen, 
    pdftitle={\myTitle},%
    pdfauthor={Ludovica de Paolis, Fabio Anselmi, Alessio Ansuini, Eugenio Piasini},%
    pdfsubject={},%
    pdfkeywords={Visual textures; Cognitive neuroscience; Convolutional neural networks; Texture synthesis; Texture perception; Brain-Score},%
    pdfcreator={pdfLaTeX},%
    pdfproducer={LaTeX with hyperref},%
    pdfdisplaydoctitle=true%
}

\title{\myTitle}
\date{} 
\author{
Ludovica de Paolis\textsuperscript{1} ({\href{mailto:ldepaoli@sissa.it} {ldepaoli@sissa.it}}),
Fabio Anselmi\textsuperscript{2} ({\href{mailto:fabio.anselmi@units.it}
{fabio.anselmi@units.it}}),
Alessio Ansuini\textsuperscript{3} ({\href{mailto:alessio.ansuini@areasciencepark.it}
{alessio.ansuini@areasciencepark.it}}),
Eugenio Piasini\textsuperscript{1} ({\href{mailto:epiasini@sissa.it} {epiasini@sissa.it}})\\

\textsuperscript{1}Neuroscience area, International School for Advanced Studies (SISSA), Trieste, Italy\\
\textsuperscript{2}Department of Mathematics, Informatics and Geosciences, Università degli Studi di Trieste, Trieste, Italy\\
\textsuperscript{3}
Department of Data Engineering, Area Science Park, Trieste, Italy
}
\date{}

\begin{document}

\vspace{-0.2cm}
\maketitle

\begingroup
\renewcommand{\thefootnote}{}
\footnotetext{Correspondence to: Ludovica de Paolis (ldepaoli@sissa.it) and Eugenio Piasini (epiasini@sissa.it)}
\footnotetext{Code available at: \url{https://github.com/ludovicadepaolis01/perceptual_misalignment}}
\endgroup

\section{Abstract}

Mathematical modeling of visual textures traces back to Julesz’s intuition that
texture perception in humans is based on local correlations between image features~\parencite{juleszTextonsElementsTexture1981}.
An influential approach for texture analysis and generation generalizes this notion to linear
correlations between the nonlinear features computed by convolutional neural networks
(CNNs), compiled into Gram matrices \parencite{gatys}. Given that CNNs are often used as models for
the visual system, it is natural to ask whether such “texture representations” spontaneously
align with the textures’ perceptual content, and in particular whether those CNNs that
are regarded as better models for the visual system also possess more human-like texture representations. Here we quantify the perceptual content captured by feature
correlations computed for a diverse pool of CNNs, and we compare it to the models’
perceptual alignment with the mammalian visual system as measured by Brain-Score \parencite{SchrimpfKubilius2018BrainScore}.
Surprisingly, we find that there is no connection between conventional measures of CNN
quality as a model of the visual system and its alignment with human texture perception.
We conclude that texture perception involves mechanisms that are distinct from those that are commonly modeled using approaches based on CNNs trained on object
recognition, possibly depending on the integration of contextual information.

\section{Keywords}
Textures, Convolutional Neural Networks, Brain-Score, Representational Similarity Analysis

\section{Introduction}

Visual textures hold a special status in vision neuroscience as
stimuli that afford a nontrivial degree of
controllability and complexity, unlike natural images or more
traditional parametric stimuli~\parencite{victorTexturesProbesVisual2017}. Because of this,
multiple mathematical frameworks for modeling and generating textures have
been established, with considerable success. One particularly
influential strand of theoretical, computational and empirical work
stems from the seminal view of {Julesz~\parencite{juleszVisualPatternDiscrimination1962,caelliPerceptualAnalyzersUnderlying1978,juleszTextonsElementsTexture1981}, who proposed that texture perception is controlled
by local correlations of visual features within an image. This view
was extended by later work that better formalized the notion of
textures as statistical objects~\parencite{zhuExploringTextureEnsembles2000} (but see \parencite{liuBoWCNNTwo2019} for a detailed historical overview). This study demonstrated the effectiveness of using features
based on the neuroscience of vision~\parencite{portilla} and
developed a connection with natural stimulus statistics through the
efficient coding paradigm~\parencite{tkacikLocalStatisticsNatural2010, victorLocalImageStatistics2012, victorTexturesProbesVisual2017, hermundstadVariancePredictsSalience2014, tesileanuEfficientCodingNatural2020,tesileanuEfficientProcessingNatural2022, caramellinoRatSensitivityMultipoint2021, zanonPredisposedLearnedPreferences2026}.

In recent years, following the rise in popularity of deep
convolutional neural networks (CNNs) as tools in computer vision, Gatys et al.\ proposed a
machine learning-based operationalization of Julesz's idea~\parencite{gatys}. The Gatys model effectively defined
a texture ensemble based on the Gram matrices formed by spatial
correlations between intermediate-layer activations of a CNN trained
for object recognition. Texture synthesis is performed by optimizing the image pixels via gradient descent to match the statistical structure of an exemplar texture. This is achieved by minimizing a \emph{Gram loss}, defined as the normalized L2 distance between the Gram matrices of the exemplar image and those of the synthesized image. This approach
was rapidly adopted in the budding fields of neural texture synthesis
and style transfer (see for instance \parencite{gatysImageStyleTransfer2016,liDiversifiedTextureSynthesis2017,ulyanov,johnsonPerceptualLossesRealTime2016,yuTextureMixerNetwork2019}), and was later improved under
multiple aspects: high resolution images processing~\parencite{snelgrove}, better methods for capturing long-range correlations within images~\parencite{liuTextureSynthesisConvolutional2016,sendik, bergerIncorporatingLongrangeConsistency2017}, usage of more sophisticated losses capable of
capturing stationary properties of the CNN's activations beyond the
pairwise correlations in the Gram loss \parencite{heitz, vacher}.

In parallel, and more broadly, CNNs trained for object recognition have become ubiquitous tools in vision neuroscience. Over the years, work from many labs has established these models as functional accounts of the ventral visual stream in monkeys~\parencite{yamins, cadieu, khalighrazavi, cadena2019, Schrimpf2020integrative, bashivanNeuralPopulationControl2019}
and humans~\parencite{storrsDiverseDeepNeural2021a,gucluDeepNeuralNetworks2015}, as well as the corresponding visual cortical areas
in rodents~\parencite{cadena2019neurips, nayebiMouseVisualCortex2023, muratore2022, muratore2025}. However, the strong predictive power of this approach for neural
activity is only partly mirrored by the capacity of these
artificial neural networks to accurately predict behavioral responses
to individual stimuli \parencite{, featherModelMetamersReveal2023, geirhosPartialSuccessClosing2021, wichmannAreDeepNeural2023,rajalinghamLargeScaleHighResolutionComparison2018,vinkenIncorporatingIntrinsicSuppression2020}.

Beyond object recognition, the ventral stream is also thought to be
important for texture perception. Indeed, the idea that texture
perception can be explained by reference to the neural mechanisms in
early vision goes at least as far back as 
\cite{turnerTextureDiscriminationGabor1986} and \cite{malikPreattentiveTextureDiscrimination1990}. More recent results in computational modeling and neurophysiological
evidence point generally to the involvement of
early-to-intermediate areas in the ventral stream \parencite{portilla, freemanMetamersVentralStream2011, freemanFunctionalPerceptualSignature2013, yu, okazawaGradualDevelopmentVisual2017, ziemba2016, henderson, matteucci, ziemba2024}. However,
overall, our understanding of the neural computations underpinning
texture perception is still limited compared to those enabling object
recognition. This \rev{makes CNN-based methods for texture analysis and synthesis particularly interesting from a neuroscience perspective, despite the recent emergence of approaches based on diffusion models for practical applications, e.g.~\parencite{wangInfiniteTextureTextguided2024}}. More specifically, it raises the question of  whether the existing quality
measures devised for CNN-based models of the ventral stream align with
the perceptual quality of textures generated based on the internal
features of those models. If such alignment exists, it would not only
suggest that object recognition and texture perception share a
significant amount of functional mechanisms, but also that these
mechanisms are among those that are well captured by CNNs trained on
object recognition. Conversely, the absence of such a relationship -- where increased fidelity to neural or behavioral object recognition data does not lead to improved modeling of human texture perception -- would indicate that current CNN architectures or training paradigms lack critical components.

In the present work, we considered a set of popular CNNs and we asked whether the models that are more ``brain-like'' in object
recogntion also tend to represent visual textures in a way that is
closer to human perception. As a metric of model performance on object
recognition, we used the standardized benchmarks provided by
Brain-Score~\parencite{SchrimpfKubilius2018BrainScore}. To measure alignment with human texture
perception, we evaluated the models on the images of the Describable
Texture Dataset~\parencite{cimpoiDescribingTexturesWild2014}, which are divided in classes according
to human annotations. We then assessed whether the ``texture
representations'' provided by the Gram matrices computed from the
models' feature maps (as in \parencite{gatys}) are organized in a way that
respect the classes subdivision in the dataset.

Our results show that there is no correlation between the goodness of
the Gram texture representation -- with respect to the clusters formed
by human-annotated labels -- and Brain-Score. This suggests that the
features governing the utility of artificial neural networks as models
of visual cortex in an object recognition context are not fully
aligned with those that can make them good models for texture
processing, therefore highlighting an underexplored path for the improvement of machine learning-based models for texture analysis and synthesis.

\section{Methods}

\subsection{The ``Gram representation'' for textures}

Our study builds on the texture analysis and synthesis method introduced by Gatys et al.~\parencite{gatys}. This approach models textures using feature representations extracted from a CNN, which are shown to support the generation of perceptually compelling texture samples. In the original implementation, given a natural texture image \(x\), the algorithm starts by feeding \(x\) as input to the network and extracting the corresponding activation maps from a predetermined set of layers. For each selected layer \(l\), the feature activations \(F^l\) are used to compute a Gram matrix \(G^l\), defined as the inner product between vector representations of the feature maps:
\begin{equation}
G^{l}_{ij} = \sum_{m} F^{l}_{im} F^{l}_{jm},
\end{equation}
where the index \(m\) runs over spatial samples of the feature maps, and \(i\) and \(j\) label feature channels at layer \(l\). This Gram matrix summarizes the correlations between feature channels, discarding spatial information and yielding a statistical description of textures. \rev{This ``Gram representation'' are thought to capture the elements of the input image that are important for texture perception. In this work, we study the Gram representations generated by a number of different CNNs.}

\subsection{Texture synthesis}
In \cite{gatys}, texture synthesis is performed by iteratively updating the pixels of an initially random image \(\hat{x}\), where pixel values are drawn from a Gaussian distribution \(\hat{x} \sim \mathcal{N}(0,1)\). This process leads to match the Gram matrices of \(\hat{x}\) to those of the target image \(x\) by minimizing the \emph{Gram loss}, defined as a weighted sum of \rev{squared L2 distances between} corresponding Gram matrices across layers:
\begin{equation}
\mathcal{L}(\hat{x}, x) = \sum_{l} \frac{1}{4M_lN_l^2} \, \lVert \hat{G}^{l} - G^{l} \rVert^2_2.
\end{equation}
\rev{where $N_l$ is the number of features in layer $l$ and $M_l$ is the number of units within each feature (note that the term $1/4M_lN_l^2$ that we use here to weight the contribution from each individual layer is not identical to the one in \cite{gatys}; see the Appendix for details on our choice of loss weighting for image synthesis). To minimize the Gram loss we used L-BFGS as in \cite{gatys}, using the implementation of the optimization algorithm provided by PyTorch~\parencite{anselPyTorch2Faster2024}.}

Minimizing this objective in pixel space generates novel images whose texture statistics, as captured by feature correlations in a convolutional neural network, match those of the original image. Importantly, while spatial structure is not preserved, the resulting images retain perceptually salient texture properties, highlighting the role of feature-level statistics in texture representation.

\rev{\subsection{CNN Layer selection}}
\rev{We implemented the texture representation and the synthesis algorithm described above using PyTorch~\parencite{anselPyTorch2Faster2024}, for a range of CNN architectures trained for object classification with ImageNet-1K. The pre-trained networks were sourced from TorchVision~\parencite{TorchVision_maintainers_and_contributors_TorchVision_PyTorch_s_Computer_2016}. The convolutional layers of the CNN originally used by \cite{gatys} (VGG-19) have five different spatial scales, corresponding to the five consecutive ``convolve-and-pool'' blocks of that architecture. \cite{gatys} showed that selecting just one layer for each of these spatial scales was enough to produce perceptually satisfying synthetic textures. Therefore,} to ensure that the results of our analysis are comparable across different networks, we extracted feature maps from five  layers from each architecture, \rev{selected in such a way that they are located at comparable relative depth across architectures; see Appendix, Supplementary Tables~\ref{tab:extr_layers},~\ref{tab:extr_layers_2} for details.}

\begin{table}[htbp]
\centering
\caption{Convolutional Neural Network architectures grouped by family.}
\label{tab:cnn_families}
\vspace{0.5em}
\small
\begin{tabular}{|c|c|c|}
\hline
\textbf{Family} & \textbf{Model} \\
\hline

\multirow{1}{*}{}
& AlexNet~\parencite{krizhevsky}
\\
\hline

\multirow{3}{*}{DenseNet~\parencite{huang}}
& DenseNet-121
\\

& DenseNet-169
\\

& DenseNet-201
\\
\hline

\multirow{1}{*}{InceptionNet~\parencite{howard}}
& Inception-v3~\parencite{szegedy2}
\\
\hline

\multirow{5}{*}{ResNet~\parencite{he}} 
  & ResNet18 
  \\
  & ResNet34 
  \\
  & ResNet50 
  \\
  & ResNet101 
  \\
  & ResNet152
  \\
  \hline

\multirow{2}{*}{VGG~\parencite{simonyan}} 
  & VGG-16
  \\
  & VGG-19 
  \\ 
  \hline

\end{tabular}
\label{tab:cnn_models}
\end{table}











\subsection{Stimulus dataset}
The Describable Textures Dataset (DTD) \parencite{cimpoiDescribingTexturesWild2014} is our input dataset of choice. It consists of 5640 images of naturalistic visual textures collected from the Internet, spanning a wide range of materials and surface appearances. The dataset is organized into 47 texture classes, each one containing 120 images. The images size ranges between 300x300px and 640x640px. We resized (scaled) every image to 224x224px, in order to match the CNNs' input spatial dimensions and to maintain a stable representation of the textures.

The class labels arise from a human annotation procedure based on the \emph{Texture Lexicon}, a psycholinguistic study introduced in \cite{bhushanTextureLexiconUnderstanding1997} with the goal of constructing a linguistically grounded vocabulary for describing visual textures. The Texture Lexicon identifies a set of texture-related adjectives commonly used by human observers (e.g., \emph{blotchy}, \emph{striped}, \emph{bumpy}, \emph{woven}) and organizes them according to perceptual relevance.

To construct DTD, the authors selected 47 representative texture terms from the Texture Lexicon and used them as labels in a large-scale annotation task involving human participants. Annotators were presented with images and asked to assign one or more texture attributes that best described the visual appearance of the image. Final class labels were determined by aggregating human responses and retaining images for which there was sufficient agreement across annotators. As a result, DTD encodes texture categories that reflect shared perceptual judgments rather than low-level image statistics or semantic object labels.

This annotation procedure makes DTD particularly suitable for studies of texture perception, as it provides a direct link between image-level texture statistics and perceptually meaningful texture descriptors grounded in human language and cognition. DTD treats textures as entities that can be categorized in a manner analogous to object images, effectively providing human-defined clusters of visual textures with a balanced distribution of images across classes. The richness and diversity of DTD thus constitute a perceptual and linguistic benchmark against which we can evaluate whether texture representations derived from CNNs in combination with the Gatys method are sufficient to preserve class-specific texture distinctions as defined by human perception.

\subsection{Clustering}
We performed Representation Similarity Analysis (RSA) \parencite{kriegeskorte} on the texture representations extracted with the Gram matrices on our pool of 12 CNNs. 
RSA was introduced in neuroscience as a method to compare object representations obtained with different means (e.g., functional Magnetic Resonance Imaging vs. Magnetoencephalography), or to compare biological and computational representations (e.g., brain activity vs. CNNs). It relies on Representational Dissimilarity Matrices (RDMs), where each entry is the pairwise dissimilarity between responses to two stimuli. We employed RSA to verify whether the RDMs produced with Gram matrices extracted from DTD show a clustering pattern consistent with the original 47 classes. For each one of the 12 CNNs, we obtained 5 RDMs -- one for each of the convolutional layers analyzed -- of dimensionality 5640x5640, where each entry is the distance among Gram matrices for one pair of images in DTD.

We tested whether Gram matrix–based texture representations capture the structure of the 47 texture classes in DTD by performing hierarchical clustering with Ward linkage function based on Euclidean distance, as implemented in scikit-learn \parencite{scikit-learn}. Hierarchical clustering is an unsupervised technique that groups data points based on their similarity, producing a tree-like representation in which clusters are progressively merged according to their pairwise distances. This hierarchical organization provides insight into the intrinsic structure of the data without relying on class labels.

To evaluate the extent to which the learned texture representations recover the human-defined categories in DTD, we treated the dataset labels as a reference standard. Specifically, we fixed the number of clusters to \(k=47\), matching the number of texture classes in the original dataset, and allowed the clustering algorithm to organize the representations accordingly. This procedure enables a direct comparison between the automatically discovered clusters and the perceptual texture classes defined by human annotations.

\subsection{Measuring clustering quality}
We quantified texture information carried in the representations within Gram matrices with Mutual Information (MI)~\parencite{mackayInformationTheoryInference2003}. In our analysis MI was computed between the real 47 classes in DTD (\textit{real classes}) and the 47 clusters found by hierarchical clustering (\textit{found clusters}). If $X$ and $Y$ are random variables representing respectively the true label of an image (\emph{real class}) and the label assigned to it by the clustering procedure (\emph{found cluster}), \rev{MI is the reduction in uncertainty, quantified as entropy, in one of the two variables associated to learning the value of other:}
\[
\operatorname{MI}(X:Y) = H(X) - H(X|Y) = H(Y) - H(Y|X) = \sum_{x,y} p(x,y)\,\log\!\frac{p(x,y)}{p(x)\,p(y)}
\]
\rev{By construction, $\operatorname{MI}(X:Y)$ is upper-bounded by the minimum between $H(X)$ and $H(Y)$. This can be seen by the expression above keeping into account that $H(X|Y)$ and $H(Y|X)$ are always nonnegative. Therefore, considering that \(Y\) follows an unknown distribution where \textit{found clusters} result from hierarchical clustering, and \(X\) is distributed uniformly among the 47 \textit{real classes} of DTD, in our case the MI is upper bounded by the entropy of $X$:}
\begin{equation}
\label{eq:MI_max}
\operatorname{MI}_{\max} = H(X) = -\sum_{i=1}^{47} p_{i}\log_{2}p_{i} = \log_{2}47 \approx 5.5 \text{bits}
\end{equation}
All empirical estimates for mutual information were corrected for potential undersampling bias using the NSB method~\parencite{Nemenman2002EntropyInferenceRevisited}, as implemented by the ndd Python package~\parencite{marsili_ndd}.

We quantified MI between \textit{real classes} and \textit{found clusters} for the \textit{found classes} distribution drawn from each of the 5 layers of the 12 CNNs. Therefore, for each model we obtained 5 MI values. 

\subsection{Brain-Score correlation}
\label{sec:brain-score-corr}
To assess the relationship between the goodness of a CNN as a model for biological object recognition and texture perception, we correlated the best MI value (i.e. the highest out of the 5 MI values computed for each model) of each CNN to Brain-Score. Brain-Score is a composite of multiple behavioral and neural benchmarks that determine how similar artificial neural networks are to the object recognition mechanisms in the human and macaque brain~\parencite{SchrimpfKubilius2018BrainScore}.  In this study, we focus on the following Brain-Score components: \emph{neural vision}, \emph{behavior vision}, and \emph{average vision}, as well as the region-specific neural predictability scores for \textit{V1, V2, V4}, and \textit{IT}. The \textit{neural vision} score summarizes a model’s ability to predict neural responses across areas of the primate ventral visual stream, while \textit{behavior vision} quantifies the alignment between model predictions and human behavioral performance on object recognition tasks. The \textit{average vision} score provides an overall summary across neural and behavioral benchmarks. The region-specific scores (\textit{V1, V2, V4, IT}) evaluate how well model representations predict neural activity in individual cortical areas involved in the hierarchical processing of visual information along the ventral stream. For each of the 12 models, we correlated the highest MI with the aforementioned Brain-Score values which we report in Appendix Supplementary Table~\ref{tab:brain_score}. We run a linear correlation with Pearson's \textit{p} and set \(\alpha=0.05\).

\section{Results}
\subsection{\rev{Representational Similarity Analysis}}

\begin{figure}[htp]
\centering
\begin{tabular}{c c c}
   \includegraphics[width=\rdmwidth, height=\rdmheight]{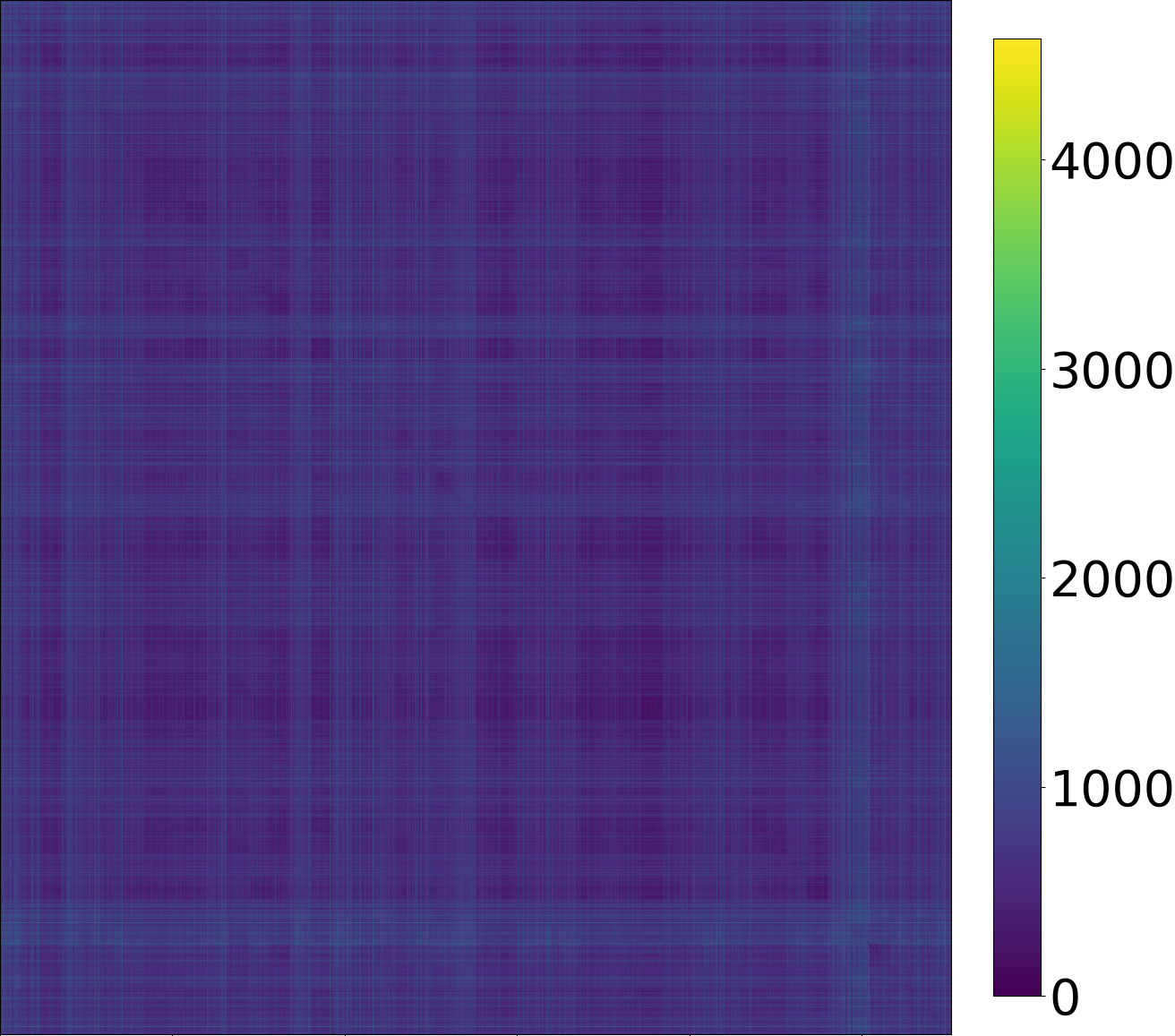} &
   \includegraphics[width=\rdmwidth, height=\rdmheight]{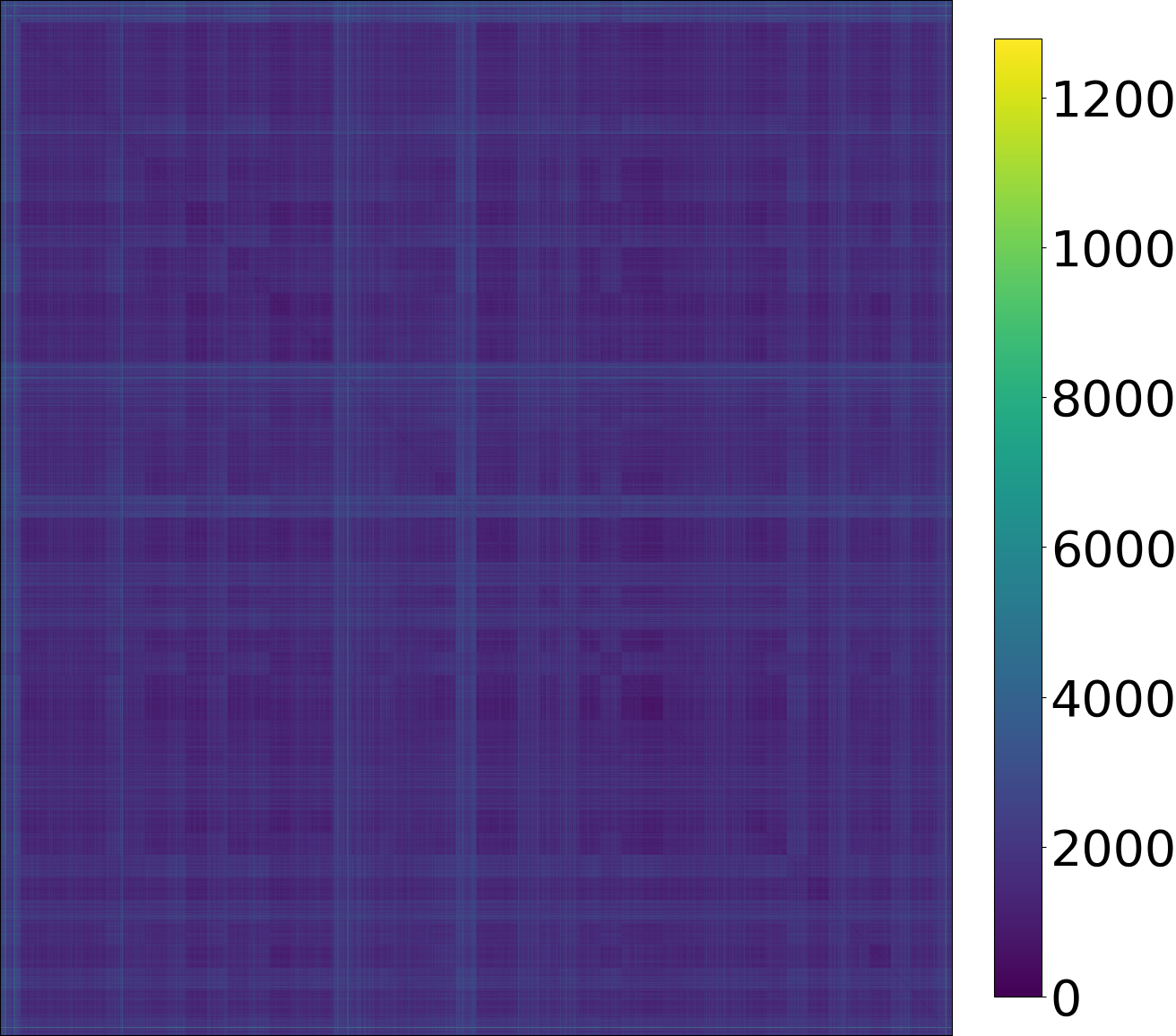} &
   \includegraphics[width=\rdmwidth, height=\rdmheight]{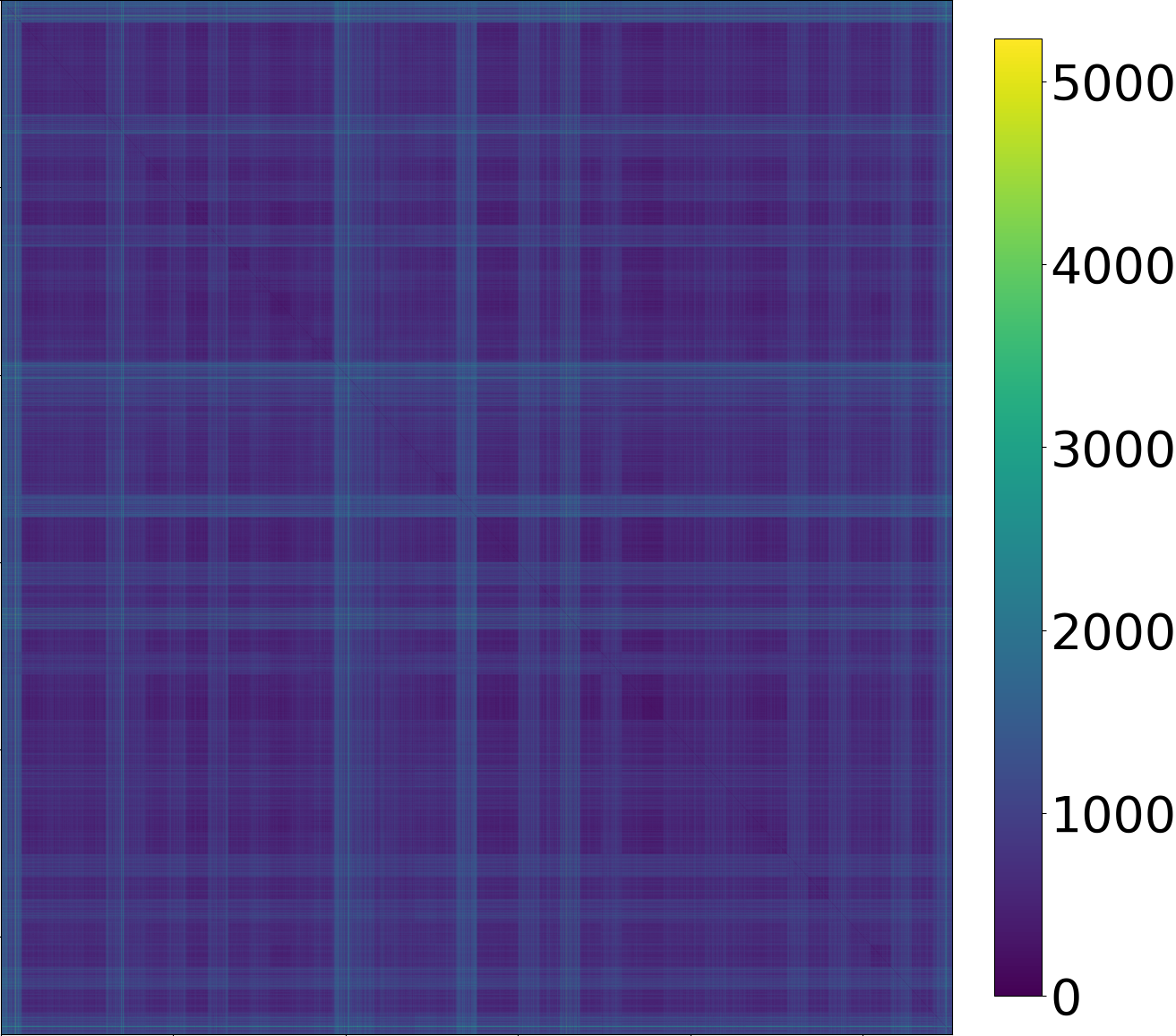} \\[-0.1em]
   Extracted layer n.\ 1 & Extracted layer n.\ 2 & Extracted layer n.\ 3
\end{tabular}

\vspace{0.5em}

\begin{tabular}{c c}
    \includegraphics[width=\rdmwidth, height=\rdmheight]{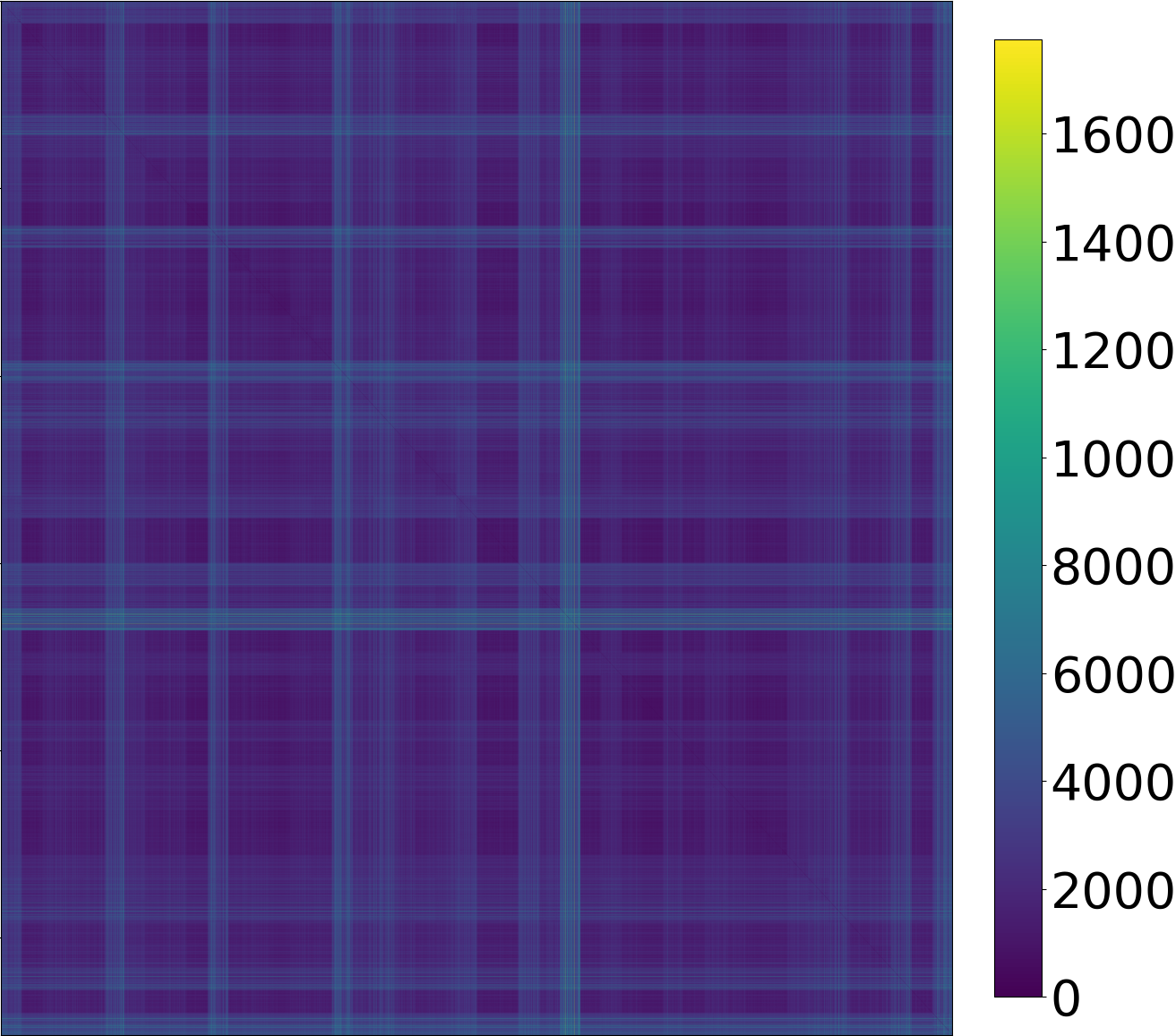} &
    \includegraphics[width=\rdmwidth, height=\rdmheight]{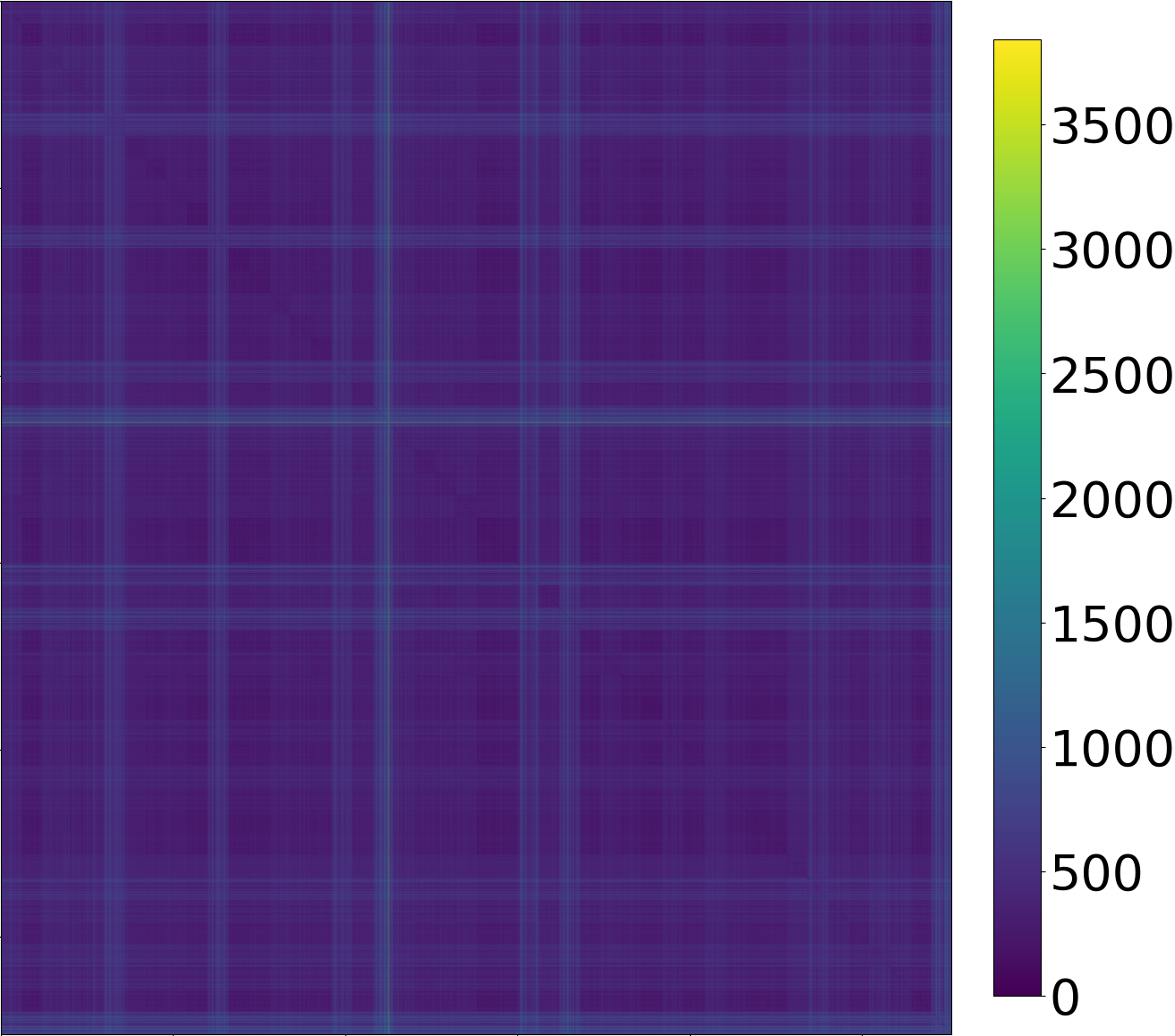}
    \\[-0.1em]
   Extracted layer n.\ 4 & Extracted layer n.\ 5 
\end{tabular}

\caption{\rev{Representational Dissimilarity Matrices obtained from the five layers analyzed for VGG-19. The size of each matrix is \(5640\times5640\), and each entry corresponds to the Euclidean distance -- relative to each matrix, as in the colorbars -- between the Gram matrix representation of one pair of images in DTD. The entries are sorted such that images belonging to each of the 47 ground-truth categories are placed next to each other.}}
\label{fig:rdms_vgg19}
\end{figure}

\rev{To understand how texture information is represented across the layers of our CNNs, we started by performing Representational Similarity Analysis. Figure~\ref{fig:rdms_vgg19} shows Representational Dissimilarity matrices (RDMs) calculated on pairs of Gram matrix representations of DTD images extracted from VGG-19. Performing the same analysis on other networks led to qualitatively similar results (see \cite{de_paolis_2025_19557466} for high-resolution RDM plots for all the networks considered in this work). Visual inspection reveals that the structure of the RDMs reflects the existence of at least some of the classes in DTD, and that for this network the clearest structure emerges in the last three layers. To understand these trends more quantitatively, we next moved to performing clustering analysis of the Gram representations.}

\subsection{Clustering analysis}
\begin{figure}
\centering
\includegraphics[width=10cm]{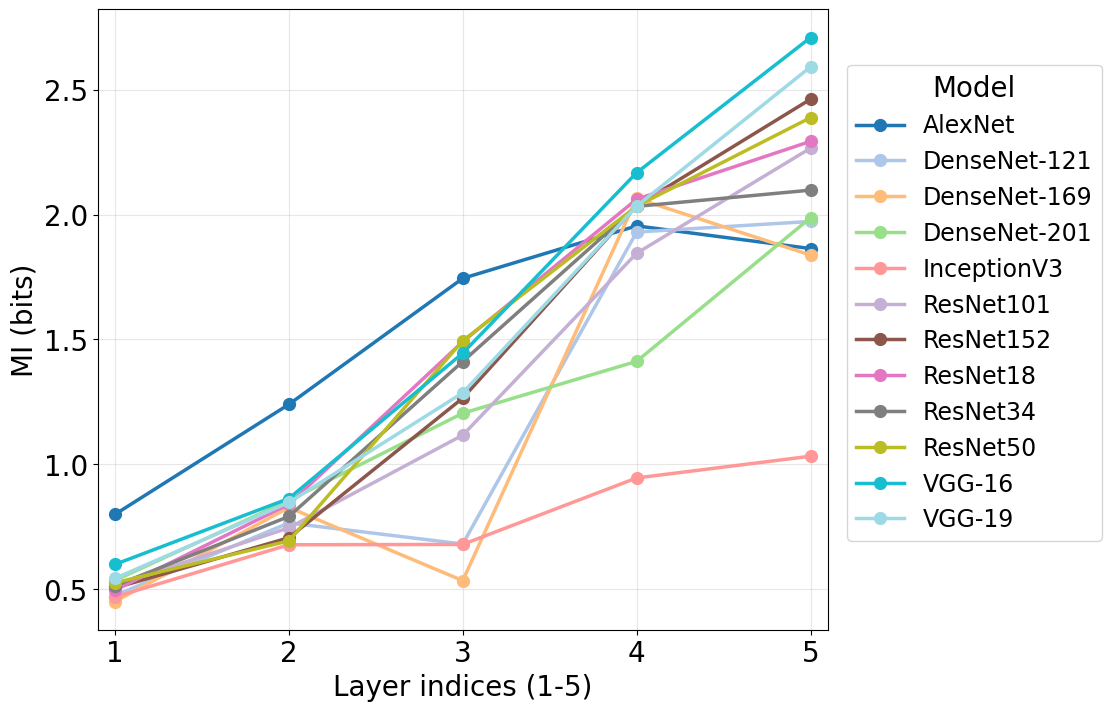} 
\caption{MI values across CNN layers identified by index (1-5). Each line corresponds to one of the 12 models as color-coded in the legend.}
\label{fig:mi_per_model_layer}
\end{figure}

Figure~\ref{fig:mi_per_model_layer} represents the values of MI by model and by layer found with hierarchical clustering. The plot shows a trend: MI grows along with the depth of the extracted features, consistently with the trend found by visual inspection in the RSA plots. The highest MI is yielded by the last layer for most models, but it does not go beyond \(\approx{2.7}\) bits, which is about half of the maximum achievable MI, according to Equation~\ref{eq:MI_max}. These numbers show that increasing the depth of the extracted representations leads to an increasing clustering quality.

The results of the correlation among the highest MI value per model and selected Brain-Score benchmarks are reported in Table~\ref{tab:correlation}. We did not find any significant correlation between any Brain-Score value and the best MI found in any model, as shown in Figure~\ref{fig:correlation_all}.

\begin{table}
\centering
\caption{Correlation between texture representation quality, expressed as MI between the clusters in the Gram representation of textures and the ground-truth labels in DTD, and quality of the networks as models of the ventral visual stream, measured by Brain-Score. For more details on Brain-Score see Section \ref{sec:brain-score-corr}.}
\vspace{0.5em}  
\resizebox{0.50\textwidth}{!}{
\begin{tabular}{|>{\raggedright\arraybackslash}m{4cm}|c|c|}
\hline
\textbf{Brain-Score metric} & \textbf{Pearson's \textit{r}} & \textbf{p-value} \\ \hline
\textbf{average vision} & -0.130 & 0.68 \\ \hline
\textbf{neural vision}  & -0.165 & 0.60 \\ \hline
\textbf{behavior vision}& -0.029 & 0.92 \\ \hline
\textbf{V1}             & -0.229 & 0.47 \\ \hline
\textbf{V2}             & -0.262 & 0.41 \\ \hline
\textbf{V4}             & -0.063 & 0.84 \\ \hline
\textbf{IT}             & -0.104 & 0.74 \\ \hline
\end{tabular}%
}
\label{tab:correlation}
\end{table}

\begin{figure}
\centering
\includegraphics[width=17cm]{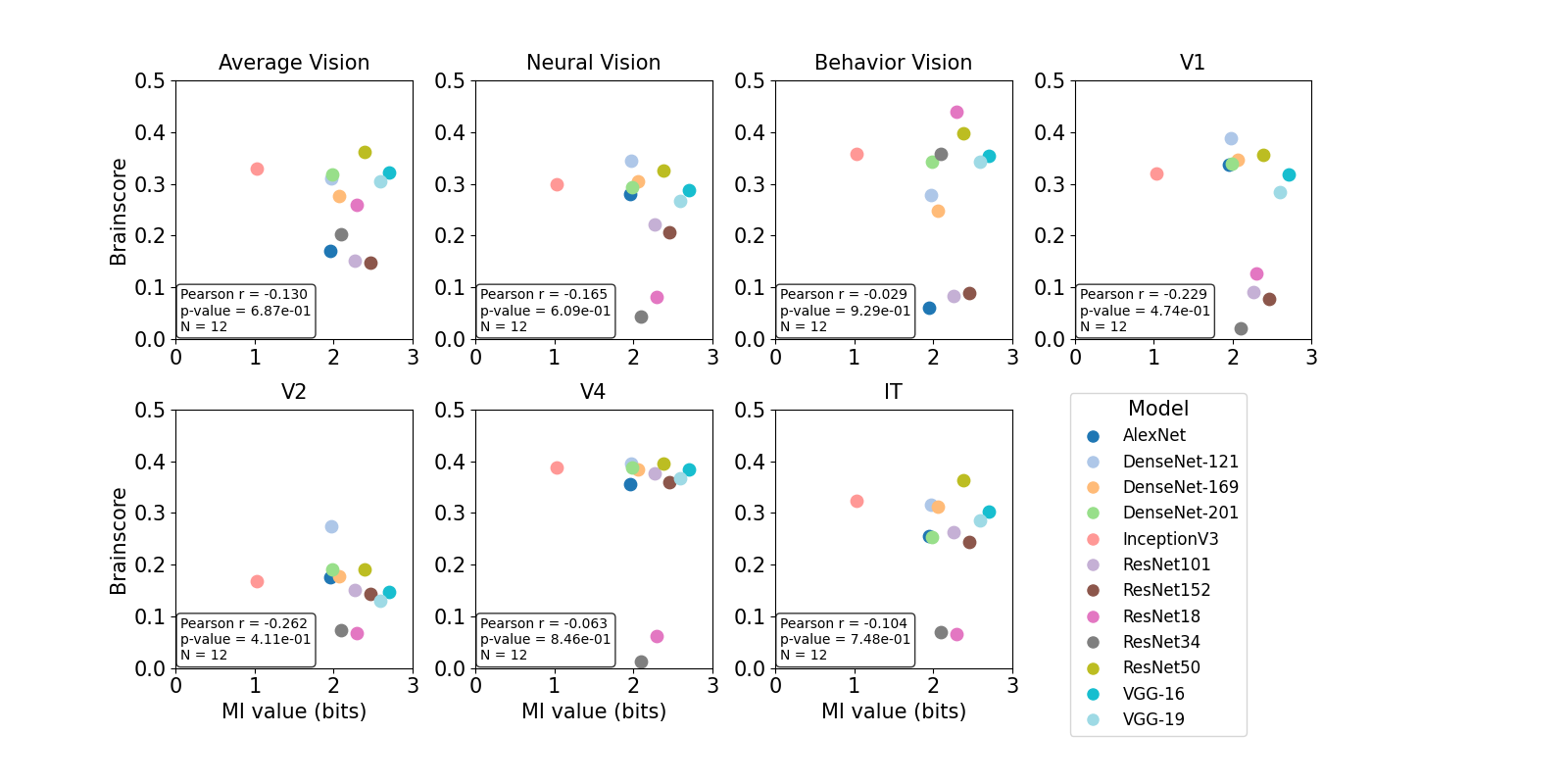} \\
\caption{Plots of the correlations with Pearson's \textit{p}. Each subplot represents the correlation of Brain-Score benchmarks \textit{average vision, neural vision, behavior vision, V1, V2, V4, IT} respectively vs. the layer with the highest MI for each of the 12 CNNs. Each color identifies a model as reported in the legend.}
\label{fig:correlation_all}
\end{figure}

\subsection{Texture synthesis}
In Figure~\ref{fig:synthesized}  we report samples of synthesized textures for a subset of selected models. The purpose of this plot is to illustrate the output of the Gatys algorithm, as implemented by us with our pool of CNNs; we underline that the results of the quantitative analyses described in the previous sections do not depend directly from this visual outcome. Each sample belongs to 4 separate DTD classes: \textit{blotchy, striped, matted, scaly}. We observe how some models achieve a better perceptual output in terms of synthesized textures, despite running the same algorithm.

\begin{table}
\begin{center}
\begin{tabular}{>{\centering\arraybackslash}m{3cm}
  >{\centering\arraybackslash}m{2cm} 
  >{\centering\arraybackslash}m{2cm}
  >{\centering\arraybackslash}m{2cm}
  >{\centering\arraybackslash}m{2cm}
  >{\centering\arraybackslash}m{2cm}
  >{\centering\arraybackslash}m{2cm}
  }
 & Blotchy & Striped & Matted & Scaly & Pebbles \\ \vspace{-0.18cm} 
Original image &
\includegraphics[width=\textureimgs, height=\textureimgs]{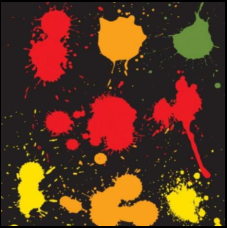} &
\includegraphics[width=\textureimgs, height=\textureimgs]{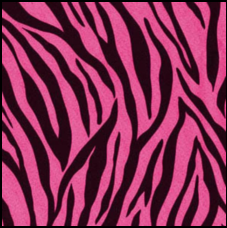} &
\includegraphics[width=\textureimgs, height=\textureimgs]{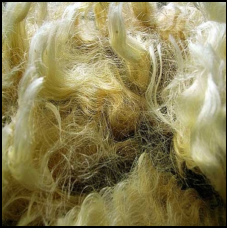} &
\includegraphics[width=\textureimgs, height=\textureimgs]{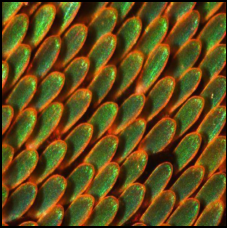} &
\includegraphics[width=\textureimgs, height=\textureimgs]{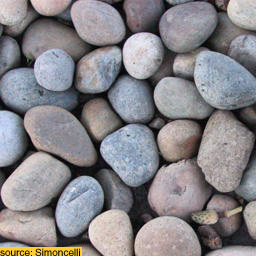}
\\ \vspace{-0.18cm}
AlexNet &
\includegraphics[width=\textureimgs, height=\textureimgs]{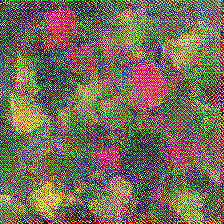} &
\includegraphics[width=\textureimgs, height=\textureimgs]{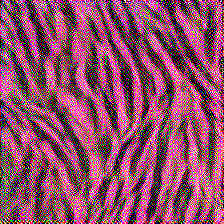} &
\includegraphics[width=\textureimgs, height=\textureimgs]{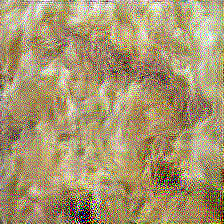} &
\includegraphics[width=\textureimgs, height=\textureimgs]{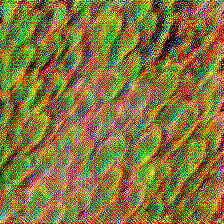} &
\includegraphics[width=\textureimgs, height=\textureimgs]{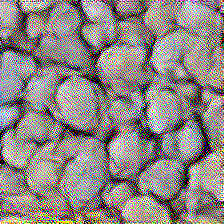}
\\ \vspace{-0.18cm}
DenseNet-169 &
\includegraphics[width=\textureimgs, height=\textureimgs]{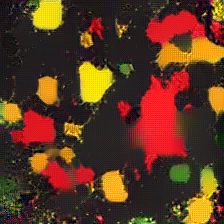} &
\includegraphics[width=\textureimgs, height=\textureimgs]{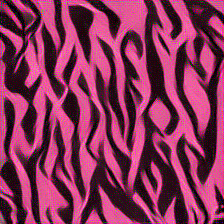} &
\includegraphics[width=\textureimgs, height=\textureimgs]{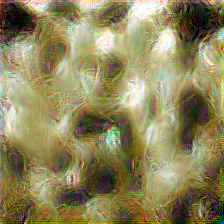} &
\includegraphics[width=\textureimgs, height=\textureimgs]{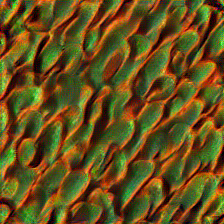} &
\includegraphics[width=\textureimgs, height=\textureimgs]{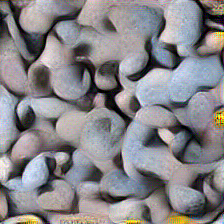}
\\ \vspace{-0.18cm}
ResNet18 &
\includegraphics[width=\textureimgs, height=\textureimgs]{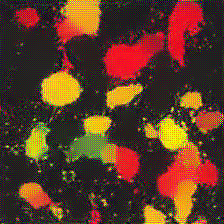} &
\includegraphics[width=\textureimgs, height=\textureimgs]{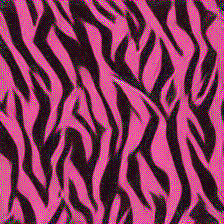} &
\includegraphics[width=\textureimgs, height=\textureimgs]{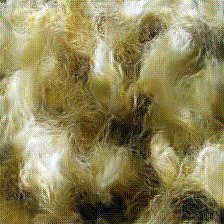} &
\includegraphics[width=\textureimgs, height=\textureimgs]{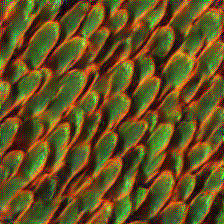} &
\includegraphics[width=\textureimgs, height=\textureimgs]{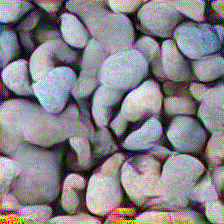}
\\ \vspace{-0.18cm}
ResNet101 &
\includegraphics[width=\textureimgs, height=\textureimgs]{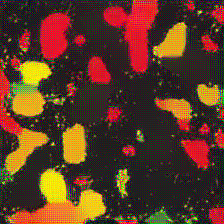} &
\includegraphics[width=\textureimgs, height=\textureimgs]{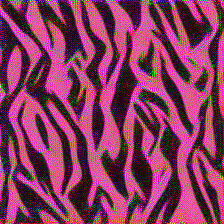} &
\includegraphics[width=\textureimgs, height=\textureimgs]{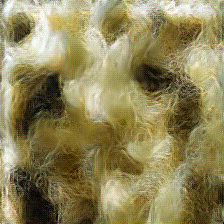} &
\includegraphics[width=\textureimgs, height=\textureimgs]{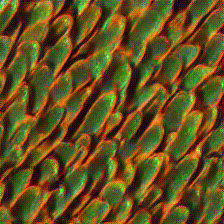} &
\includegraphics[width=\textureimgs, height=\textureimgs]{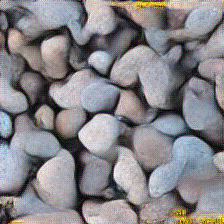} 
\\ \vspace{-0.18cm}
VGG-19 &
\includegraphics[width=\textureimgs, height=\textureimgs]{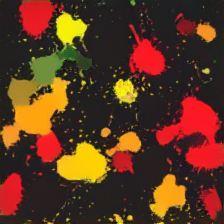} &
\includegraphics[width=\textureimgs, height=\textureimgs]{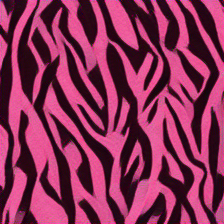} &
\includegraphics[width=\textureimgs, height=\textureimgs]{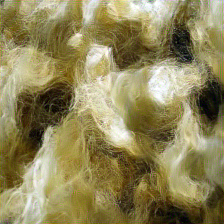} &
\includegraphics[width=\textureimgs, height=\textureimgs]{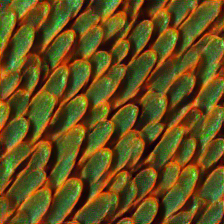} &
\includegraphics[width=\textureimgs, height=\textureimgs]{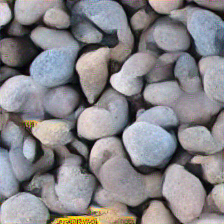}
\\ 
\end{tabular}
\end{center}

\captionof{figure}{Textures generated with the Gatys algorithm applied to a subsample of CNNs (rows). The 4 original images (top row) belong to 4 different classes in DTD (columns: \emph{blotchy; striped; matted; scaly}). The last column represents one of the images originally used by Gatys et al., which we synthesized as a reference. Textures generated by the whole pool of 12 models can be found in Appendix (Supplementary Figures \ref{fig:synthesized_all_1}, \ref{fig:synthesized_all_2})}
\label{fig:synthesized}
\end{table}

\section{Discussion}
In this study, we have investigated the quality of the Gram texture representation proposed by \cite{gatys} in terms of its agreement with the categorical grouping in the Describable Texture Dataset \parencite{cimpoiDescribingTexturesWild2014}, for a diverse pool of deep convolutional networks. We have compared this measure of perceptual quality with Brain-Score, an independent benchmark for ``brain-like'' operation of the CNNs in object recognition settings \parencite{SchrimpfKubilius2018BrainScore,Schrimpf2020integrative}.

Our analyses lead to three main conclusions. First, the perceptual quality of the Gram representations tends, for most networks, to grow monotonically with the depth of the layers analyzed (Figure~\ref{fig:mi_per_model_layer}). This is generally compatible with recent results suggesting that texture processing is distributed along intermediate areas of the ventral visual stream inclusing V2 and V4~\parencite{ziemba2024, okazawaGradualDevelopmentVisual2017} and possibly IT~\parencite{zhivagoTextureDiscriminabilityMonkey2014}.

Second, the Gram representation is unable to recover the full class structure of the data, as determined by the human-assigned labels. Quantitatively, this is seen by the value of the clustering MI reaching only about half of the theoretical maximum (Figure~\ref{fig:mi_per_model_layer}). 
This highlights that, despite its popularity, the Gram representation is far from perfect. In further work, we aim at exploring whether the perceptual improvements associated with more sophisticated versions of the representation translate to significant improvements in MI on the DTD.

Third, the quality of the CNNs as implicit models for texture perception is not correlated with their quality as models of the ventral stream in an object recognition setting (Figure~\ref{fig:correlation_all}). This is despite the fact that there is substantial variability in both metrics, as is well known for Brain-Score, and as can be also appreciated visually for the textures in Figure~\ref{fig:synthesized}. It is not simply due for instance to all models having roughly the same texture representation capacity. This result is surprising, because one could have expected that the functional features captured by Brain-Score are general enough to at least partly include those that are necessary to support human-like texture perception. Instead, this appears not to be the case. We conclude that in order to design methods of texture analysis and synthesis that are better aligned with human perception we need to look beyond standard CNNs trained for object recognition, for instance by incorporating unsupervised or self-supervised learning~\parencite{matteucci}, multimodality~\parencite{Radford2021LearningTransferableVisual} or different architectures, such as those with attention mechanisms~\parencite{Vaswani2017AttentionIsAllYouNeed, guo}.

\rev{Finally, we note that our conclusions are broadly compatible with another series of works relating human vision and texture encoding in CNNs. In an influential paper, \cite{geirhosImageNettrainedCNNsAre2019} showed that CNNs trained for object classification on ImageNet tend to classify images based on texture information more than shape information, in a way that differs markedly from human behavior. However, subsequent work within that research strand point to the idea that, in fact, shape and texture information do coexist within the representation layers of CNNs, and that this so-called ``texture bias'' emerges primarily at the level of the decision head, which under standard training conditions learns to prioritize texture information over shape information \parencite{NEURIPS2020_db5f9f42,islamShapeTextureUnderstanding2021}. Our notion of ``texture information'' (the MI between the Gram clusters and the ground-truth DTD labels) increases with layer depth across most models, generalizing the observations in \cite{NEURIPS2020_db5f9f42}, which shows (Supplemental Figure C.4) that for AlexNet the texture information increases rapidly in early convolutional layers and then flattens out towards the end of the hierarchy. This is very similar to the trend we report for AlexNet in Figure \ref{fig:mi_per_model_layer}. Another interesting observation in \cite{NEURIPS2020_db5f9f42} is that neurally-motivated architectures like CorNET (\cite{kubiliusCORnetModelingNeural2018}; designed by the same team as Brain-Score) do not fare better than standard CNNs from the point of view of their texture bias in classification. This echoes our own results showing that architectures that are better models of ventral stream processing in the canonical sense (related to object identity processing, measured by Brain-Score) are not necessarily more human-like in how they process textures.}

\section{Conclusions}
Our work highlights a number of limitations of popular machine learning-based approaches for visual texture analysis and synthesis. At the same time, it provides a simple example of a visual function that is likely performed in the ventral stream but for which the existing quality metrics of neural and behavioral fidelity for models seem to have very little predictive power. In future work, we plan to extend our analyses to different architectures and texture representations, and to develop a more direct and systematic way of testing the perceptual quality of generated textures with large-scale human psychophysics. Going forward, we hope that our approach can provide an additional guide for the development of truly general-purpose models of the mammalian visual system.




\section{Acknowledgments}
EP was partially supported by by the European Union –- NextGenerationEU –- PNRRM4C2-I.1.1, in the framework of PRIN Project no. 2022XE8X9E, CUP:G53D23004590001.
\\AA was supported by the project ``Supporto alla diagnosi di malattie rare tramite l’intelligenza artificiale'' CUP: F53C22001770002, from the project ``Valutazione automatica delle immagini diagnostiche tramite l’intelligenza artificiale'', CUP: F53C22001780002, and by the European Union – NextGenerationEU within the project PNRR ``PRP@CERIC'' IR0000028 - Mission 4 Component 2 Investment 3.1 Action 3.1.1.
\printbibliography

\pagebreak
\section{Appendix}

\rev{\subsection{Loss normalization}
\label{sec:loss-normalization}

In this section we derive the weighting rule we used to combine the Gram losses from multiple layers.

Consider the Gram matrix $G$ for a certain layer $l$. Its $(i,j)$-th entry is
\[
G_{ij} = \sum_{n=1}^M F_{i}^n F_{j}^n
\]
where the index $n$ runs over all elements of the $i$-th feature map $F_i$ (so for instance in a typical convolutional network $M$ would be the product of the height and width of the feature map). Assume that the individual activations of the network units are roughly normally distributed:
\[
F_i^n \sim \mathcal{N}(0,1)
\]
This is a reasonable assumption in presence of batch normalization. Our goal is now to understand how the ``Gram loss'' for this layer is distributed. The Gram loss is
\[
\mathcal{L}_l = \sum_{i=1}^N\sum_{j=1}^N \left(G_{ij}^{\text{rec}}- G_{ij}^{\text{orig}}\right)^2
\]
For a fixed value of $i$, $j$ and $n$ let's define two new random variables $A$ and $B$:
\[
A = \left(\frac{F_i^n + F_j^n}{\sqrt{2}}\right)^2 \quad , \quad B = \left(\frac{F_i^n - F_j^n}{\sqrt{2}}\right)^2
\]
from which we get
\[
\frac{A-B}{2} = \frac{1}{2}\left[\frac{F_i^2+F_j^2+2F_iF_j}{2} - \frac{F_i^2+F_j^2-2F_iF_j}{2}\right] \\= \frac{1}{4}\left[4F_iFj\right] = F_iF_j
\]
so we can think of $G_{ij}$ as the sum of $M$ iid terms, each of which is distributed like $(A-B)/2$.
By observing that $(F_i^n + F_j^n)/\sqrt{2}\sim\mathcal{N}(0,1)$, we see that $A$ and $B$ are each simply the square of a standard Normal. Therefore they both follow a Chi-squared distribution with one degree of freedom
\[
A\sim\chi^2_1\quad,\quad B\sim\chi^2_1
\]
which means that
\[
\mathbb{E}[A] = \mathbb{E}[B] = 1 \quad,\quad \operatorname{Var}[A]=\operatorname{Var}[B] = 2
\]
Therefore, $A-B$ is the difference of two independent random variables with mean zero and variance 2. This implies that, for any $n$,
\begin{gather*}
\mathbb{E}[2F_i^nF_j^n] = \mathbb{E}[A-B] = 0\\
\operatorname{Var}[2F_i^nF_j^n] = \operatorname{Var}[A] + \operatorname{Var}[B] = 4
\end{gather*}
and therefore
\begin{gather*}
\mathbb{E}[F_i^nF_j^n] = 0\\
\operatorname{Var}[F_i^nF_j^n] = 2
\end{gather*}
So $G_{ij}$ is the sum of a large number $M$ of iid random variables, each with mean 0 and variance 2. By the central limit theorem, then,
\[
G_{ij} \sim \mathcal{N}\left(0,\sqrt{2M}\right)
\]
If this is true for both the original and reconstructed Gram, then
\[
G_{ij}^{\text{rec}}- G_{ij}^{\text{orig}} \sim \mathcal{N}(0, 2\sqrt{M})
\]
and therefore
\[
\frac{G_{ij}^{\text{rec}}- G_{ij}^{\text{orig}}}{2\sqrt{M}} \sim \mathcal{N}(0,1)
\]

Now, consider the quantity
\[
\frac{\mathcal{L}_l}{4M} = \sum_{i=1}^N\sum_{j=1}^N \left(\frac{G_{ij}^{\text{rec}}- G_{ij}^{\text{orig}}}{2\sqrt{M}}\right)^2
\]
It is clear that $\mathcal{L}_l/4M$ is the sum of $N^2$ squared standard Normal random variables. Therefore,
\[
\frac{\mathcal{L}_l}{4M} \sim \chi^2_{N^2}
\]
and as a consequence
\[
\mathbb{E}\left[\frac{\mathcal{L}_l}{4M}\right] = N^2
\]
Finally, this implies that
\[
\mathbb{E}\left[\mathcal{L}^l\right] = 4MN^2
\]
In other words, if the activations $F_i^n$ are standardized, the expected value of the Gram loss for a layer with $N$ features of size $M$ is $4MN^2$, just as a consequence of the size and shape of the layer. Therefore, to make sure that Gram losses coming from different layers have the same weight in our total loss we should divide each layer-specific Gram loss by $4MN^2$, where $M$ and $N$ are the values specific to that layer.
\\

\subsection{Supplementary Tables}
\begin{table}[htbp]
\centering
\caption{\rev{Extracted layers from each CNN and relative Gram matrix dimensionality. The names refer to the terminology of the models' versions in Torchvision. For models whose layers are not associated with a name, we report the absolute index.}}
\label{tab:extr_layers}
\vspace{0.5em}
\small
\begin{tabular}{|l|p{8cm}|c|}

\hline
\textbf{Model} & \textbf{Extracted layers} & \textbf{Gram dimensionality} \\
\hline

\multirow{5}{*}{AlexNet}
& \multirow{5}{*}{\texttt{layer idx=[0, 3, 6, 8, 10]}}
& L0=\(64(64+1)/2=2080\)\\
& & L3=\(192(192+1)/2=18520\)\\
& & L6=\(384(384+1)/2=73920\)\\
& & L8=\(256(256+1)/2=32896\)\\
& & L10=\(256(256+1)/2=32896\)\\
\hline

\multirow{5}{*}{DenseNet-121}
& \texttt{bn\_0=features.norm0}
& bn0=\(64(64+1)/2=2080\) \\
& \texttt{bn\_1=features.denseblock1.denselayer1.norm2}
& bn1=\(128(128+1)/2=8256\) \\
& \texttt{bn\_2=features.denseblock2.denselayer1.norm2}
& bn2=\(128(128+1)/2=8256\) \\
& \texttt{bn\_3=features.denseblock3.denselayer1.norm2}
& bn3=\(128(128+1)/2=8256\) \\
& \texttt{bn\_4=features.denseblock4.denselayer1.norm2}
& bn4=\(128(128+1)/2=8256\) \\
\hline

\multirow{5}{*}{DenseNet-169}
& \texttt{bn\_0=features.norm0}
& bn0=\(64(64+1)/2=2080\) \\
& \texttt{bn\_1=features.denseblock1.denselayer1.norm2}
& bn1=\(128(128+1)/2=8256\) \\
& \texttt{bn\_2=features.denseblock2.denselayer1.norm2}
& bn2=\(128(128+1)/2=8256\) \\
& \texttt{bn\_3=features.denseblock3.denselayer1.norm2}
& bn3=\(128(128+1)/2=8256\) \\
& \texttt{bn\_4=features.denseblock4.denselayer1.norm2}
& bn4=\(128(128+1)/2=8256\) \\
\hline

\multirow{5}{*}{DenseNet-201}
& \texttt{bn\_0=features.norm0}
& bn0=\(64(64+1)/2=2080\) \\
& \texttt{bn\_1=features.denseblock1.denselayer1.norm2}
& bn1=\(128(128+1)/2=8256\) \\
& \texttt{bn\_2=features.denseblock2.denselayer1.norm2}
& bn2=\(128(128+1)/2=8256\) \\
& \texttt{bn\_3=features.denseblock3.denselayer1.norm2}
& bn3=\(128(128+1)/2=8256\) \\
& \texttt{bn\_4=features.denseblock4.denselayer1.norm2}
& bn4=\(128(128+1)/2=8256\) \\
\hline

\end{tabular}
\end{table}

\begin{table}[htbp]
\centering
\caption{(Continued from previous page)}
\label{tab:extr_layers_2}
\vspace{0.5em}
\small
\begin{tabular}{|l|p{8cm}|c|}

\hline
\textbf{Model} & \textbf{Extracted layers} & {Gram matrix dimensionality}\\
\hline

\multirow{5}{*}{Inception-v3}
& \texttt{bn\_0=Conv2d\_1a\_3x3.bn}
& bn0=\(32(32+1)/2=561\) \\
& \texttt{bn\_1=Conv2d\_2a\_3x3.bn}
& bn1=\(32(32+1)/2=561\) \\
& \texttt{bn\_2=Conv2d\_2b\_3x3.bn}
& bn2=\(64(64+1)/2=2080\) \\
& \texttt{bn\_3=Conv2d\_3b\_1x1.bn}
& bn3=\(80(80+1)/2=3240\) \\
& \texttt{bn\_4=Conv2d\_4a\_3x3.bn}
& bn4=\(192(192+1)/2=185208\) \\
\hline

\multirow{5}{*}{ResNet18}
& \texttt{bn\_0=layer1.0.bn1}
& bn0=\(64(64+1)/2=2080\) \\
& \texttt{bn\_1=layer1.0.bn1}
& bn1=\(64(64+1)/2=2080\) \\
& \texttt{bn\_2=layer2.0.bn1}
& bn2=\(128(128+1)/2=8256\) \\
& \texttt{bn\_3=layer3.0.bn1}
& bn3=\(256(256+1)/2=32896\) \\
& \texttt{bn\_4=layer4.0.bn1}
& bn4=\(512(215+1)/2=131280\) \\
\hline

\multirow{5}{*}{ResNet34}
& \texttt{bn\_0=layer1.0.bn1}
& bn0=\(64(64+1)/2=2080\) \\
& \texttt{bn\_1=layer1.0.bn1}
& bn1=\(64(64+1)/2=2080\) \\
& \texttt{bn\_2=layer2.0.bn1}
& bn2=\(128(128+1)/2=8256\) \\
& \texttt{bn\_3=layer3.0.bn1}
& bn3=\(256(256+1)/2=32896\) \\
& \texttt{bn\_4=layer4.0.bn1}
& bn4=\(512(215+1)/2=131280\) \\
\hline

\multirow{5}{*}{ResNet50}
& \texttt{bn\_0=layer1.0.bn1}
& bn0=\(64(64+1)/2=2080\) \\
& \texttt{bn\_1=layer1.0.bn1}
& bn1=\(64(64+1)/2=2080\) \\
& \texttt{bn\_2=layer2.0.bn1}
& bn2=\(128(128+1)/2=8256\) \\
& \texttt{bn\_3=layer3.0.bn1}
& bn3=\(256(256+1)/2=32896\) \\
& \texttt{bn\_4=layer4.0.bn1}
& bn4=\(512(215+1)/2=131280\) \\
\hline

\multirow{5}{*}{ResNet101}
& \texttt{bn\_0=layer1.0.bn1}
& bn0=\(64(64+1)/2=2080\) \\
& \texttt{bn\_1=layer1.0.bn1}
& bn1=\(64(64+1)/2=2080\) \\
& \texttt{bn\_2=layer2.0.bn1}
& bn2=\(128(128+1)/2=8256\) \\
& \texttt{bn\_3=layer3.0.bn1}
& bn3=\(256(256+1)/2=32896\) \\
& \texttt{bn\_4=layer4.0.bn1}
& bn4=\(512(215+1)/2=131280\) \\
\hline

\multirow{5}{*}{ResNet152}
& \texttt{bn\_0=layer1.0.bn1}
& bn0=\(64(64+1)/2=2080\) \\
& \texttt{bn\_1=layer1.0.bn1}
& bn1=\(64(64+1)/2=2080\) \\
& \texttt{bn\_2=layer2.0.bn1}
& bn2=\(128(128+1)/2=8256\) \\
& \texttt{bn\_3=layer3.0.bn1}
& bn3=\(256(256+1)/2=32896\) \\
& \texttt{bn\_4=layer4.0.bn1}
& bn4=\(512(215+1)/2=131280\) \\
\hline

\multirow{5}{*}{VGG-16}
&  \multirow{5}{*}{\texttt{bn idx=[1, 8, 15, 25, 35]}} 
& bn1=\(64(64+1)/2=2080\) \\
& & bn8=\(128(128+1)/2=8256\) \\
& & bn15=\(256(256+1)/2=32896\) \\
& & bn25=\(512(215+1)/2=131280\) \\
& & bn35=\(512(215+1)/2=131280\) \\
\hline

\multirow{5}{*}{VGG-19}
& \multirow{5}{*}{\texttt{bn idx=[1, 8, 15, 28, 41]}}
& bn1=\(64(64+1)/2=2080\) \\
& & bn8=\(128(128+1)/2=8256\) \\
& & bn15=\(256(256+1)/2=32896\) \\
& & bn25=\(512(215+1)/2=131280\) \\
& & bn35=\(512(215+1)/2=131280\) \\
\hline

\end{tabular}
\end{table}

\begin{table}
\centering
\caption{\rev{Summary of the values from Brain-Score benchmarks that we selected for our study (columns), for the 12 CNNs from our pool (rows). In the second column and second row, for models and benchmarks respectively, we report the .csv coordinates of the leaderboard. Our Brain-Score values were exported from the leaderboard website on September 8th 2025.}}
\vspace{0.5em}
\resizebox{\textwidth}{!}{%
\begin{tabular}{|*{9}{c|}}
\hline
\textbf{Model} & \textbf{Model coordinates} & \textbf{average vision} & \textbf{neural vision} & \textbf{behavior vision} & \textbf{V1} & \textbf{V2} & \textbf{V4} & \textbf{IT} \\ \hline
\textbf{Benchmark coordinates} & & \$D\$2 & \$E\$2 & \$DB\$2 & \$F\$2 & \$AM\$2 & \$AP\$2 & \$AV\$2 \\ \hline
\textbf{AlexNet} & \$B\$290 & 0.171 & 0.281 & 0.06 & 0.338 & 0.176 & 0.355 & 0.255 \\ \hline
\textbf{Densenet-121} & \$B\$128 & 0.311 & \textbf{0.344} & 0.278 & \textbf{0.388} & \textbf{0.275} & \textbf{0.396} & 0.315 \\ \hline
\textbf{DenseNet-169} & \$B\$158 & 0.276 & 0.305 & 0.248 & 0.346 & 0.178 & 0.384 & 0.312 \\ \hline
\textbf{DenseNet-201} & \$B\$94 & 0.318 & 0.293 & 0.343 & 0.34 & 0.19 & 0.388 & 0.253 \\ \hline
\textbf{Inception-v3} & \$B\$84 & 0.33 & 0.3 & 0.358 & 0.321 & 0.169 & \textbf{0.388} & 0.324 \\ \hline
\textbf{Resnet18} & \$B\$354 & 0.152 & 0.221 & 0.084 & 0.091 & 0.151 & 0.377 & 0.263 \\ \hline
\textbf{Resnet34} & \$B\$357 & 0.147 & 0.206 & 0.089 & 0.077 & 0.143 & 0.359 & 0.244 \\ \hline
\textbf{Resnet50} & \$B\$60 & \textbf{0.362} & 0.326 & \textbf{0.397} & 0.356 & 0.191 & 0.395 & \textbf{0.363} \\ \hline
\textbf{Resnet101} & \$B\$176 & 0.26 & 0.081 & 0.44 & 0.127 & 0.068 & 0.062 & 0.066 \\ \hline
\textbf{Resnet152} & \$B\$228 & 0.202 & 0.044 & 0.359 & 0.021 & 0.074 & 0.012 & 0.069 \\ \hline
\textbf{VGG-16} & \$B\$104 & 0.322 & 0.288 & 0.354 & 0.319 & 0.147 & 0.385 & 0.303 \\ \hline
\textbf{VGG-19} & \$B\$148 & 0.305 & 0.267 & 0.343 & 0.285 & 0.13 & 0.367 & 0.285 \\ \hline
\end{tabular}
}
\label{tab:brain_score}
\end{table}
\clearpage

\subsection{Supplementary Figures}
\renewcommand{\figurename}{Supplementary Figure}
\setcounter{figure}{0}}

\begin{table}
\begin{center}
\begin{tabular}{>{\centering\arraybackslash}m{3cm}
  >{\centering\arraybackslash}m{2cm} 
  >{\centering\arraybackslash}m{2cm}
  >{\centering\arraybackslash}m{2cm}
  >{\centering\arraybackslash}m{2cm}
  >{\centering\arraybackslash}m{2cm}
  >{\centering\arraybackslash}m{2cm}}
 & Blotchy & Striped & Matted & Scaly & Pebbles \\ \vspace{-0.18cm} 
Original image &
\includegraphics[width=\textureimgs, height=\textureimgs]{figures/orig_blotchy_b0_s29999.png} &
\includegraphics[width=\textureimgs, height=\textureimgs]{figures/orig_striped_b0_s29999.png} &
\includegraphics[width=\textureimgs, height=\textureimgs]{figures/orig_matted_b0_s29999.png} &
\includegraphics[width=\textureimgs, height=\textureimgs]{figures/orig_scaly_b0_s29999.png} &
\includegraphics[width=\textureimgs, height=\textureimgs]{figures/pebbles.jpg}
\\ \vspace{-0.18cm}
AlexNet &
\includegraphics[width=\textureimgs, height=\textureimgs]{figures/alexnet_blotchy_reco_s5998.png} &
\includegraphics[width=\textureimgs, height=\textureimgs] {figures/alexnet_striped_reco_s5998.png} &
\includegraphics[width=\textureimgs, height=\textureimgs]{figures/alexnet_matted_reco_s5998.png} &
\includegraphics[width=\textureimgs, height=\textureimgs]{figures/alexnet_scaly_reco_s5998.png} &
\includegraphics[width=\textureimgs, height=\textureimgs]{figures/alexnet_pebble_reco_s5998.png}
\\ \vspace{-0.18cm}
DenseNet-121 &
\includegraphics[width=\textureimgs, height=\textureimgs]{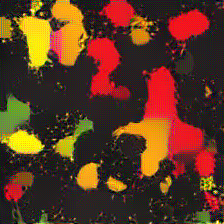} &
\includegraphics[width=\textureimgs, height=\textureimgs]{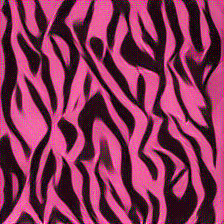} &
\includegraphics[width=\textureimgs, height=\textureimgs]{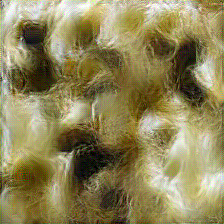} &
\includegraphics[width=\textureimgs, height=\textureimgs]{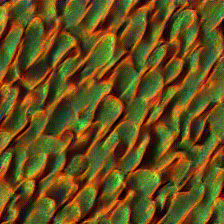} &
\includegraphics[width=\textureimgs, height=\textureimgs]{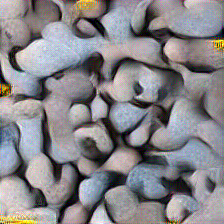}
\\ \vspace{-0.18cm}
DenseNet-169 &
\includegraphics[width=\textureimgs, height=\textureimgs]{figures/densenet169_blotchy_reco_s5998.png} &
\includegraphics[width=\textureimgs, height=\textureimgs]{figures/densenet169_striped_reco_s5998.png} &
\includegraphics[width=\textureimgs, height=\textureimgs]{figures/densenet169_matted_reco_s5998.png} &
\includegraphics[width=\textureimgs, height=\textureimgs]{figures/densenet169_scaly_reco_s5998.png} &
\includegraphics[width=\textureimgs, height=\textureimgs]{figures/densenet169_pebble_reco_s5998.png}
\\ \vspace{-0.18cm}
DenseNet-201 &
\includegraphics[width=\textureimgs, height=\textureimgs]{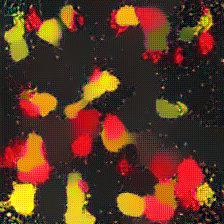} &
\includegraphics[width=\textureimgs, height=\textureimgs]{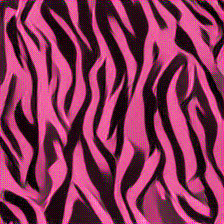} &
\includegraphics[width=\textureimgs, height=\textureimgs]{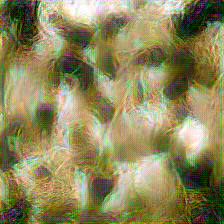} &
\includegraphics[width=\textureimgs, height=\textureimgs]{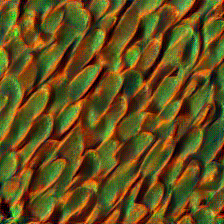} &
\includegraphics[width=\textureimgs, height=\textureimgs]{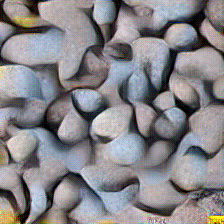}
\\ \vspace{-0.18cm}
Inception-v3 &
\includegraphics[width=\textureimgs, height=\textureimgs]{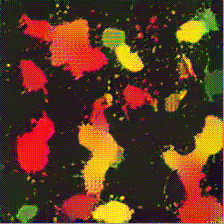} &
\includegraphics[width=\textureimgs, height=\textureimgs]{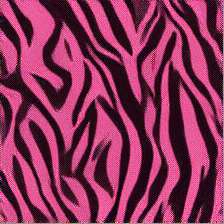} &
\includegraphics[width=\textureimgs, height=\textureimgs]{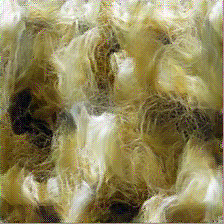} &
\includegraphics[width=\textureimgs, height=\textureimgs]{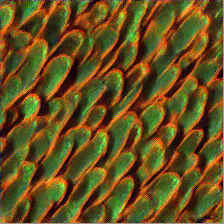} &
\includegraphics[width=\textureimgs, height=\textureimgs]{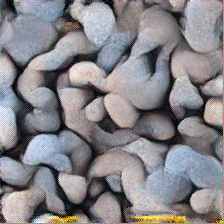}
\\ \vspace{-0.18cm}
ResNet18 &
\includegraphics[width=\textureimgs, height=\textureimgs]{figures/resnet18_blotchy_reco_s5998.png} &
\includegraphics[width=\textureimgs, height=\textureimgs]{figures/resnet18_striped_reco_s5998.png} &
\includegraphics[width=\textureimgs, height=\textureimgs]{figures/resnet18_matted_reco_s5998.png} &
\includegraphics[width=\textureimgs, height=\textureimgs]{figures/resnet18_scaly_reco_s5998.png} &
\includegraphics[width=\textureimgs, height=\textureimgs]{figures/resnet18_pebble_reco_s5998.png}
\\ \vspace{-0.18cm}
\end{tabular}
\end{center}
\captionof{figure}{Textures generated with the Gatys algorithm with the 12 CNNs of choice
(rows). The 4 original images (top row) belong to 4 different classes in DTD (columns: \emph{blotchy;
striped; matted; scaly}). The last column represents one of the images originally used by Gatys et al., which we synthesized as a reference.}
\label{fig:synthesized_all_1}
\end{table}

\begin{table}
\begin{center}
\begin{tabular}{>{\centering\arraybackslash}m{3cm}
  >{\centering\arraybackslash}m{2cm} 
  >{\centering\arraybackslash}m{2cm}
  >{\centering\arraybackslash}m{2cm}
  >{\centering\arraybackslash}m{2cm}
  >{\centering\arraybackslash}m{2cm}
  >{\centering\arraybackslash}m{2cm}}
 & Blotchy & Striped & Matted & Scaly & Pebbles\\  \vspace{-0.18cm}
 
ResNet34 &
\includegraphics[width=\textureimgs, height=\textureimgs]{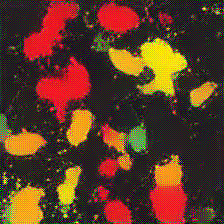} &
\includegraphics[width=\textureimgs, height=\textureimgs]{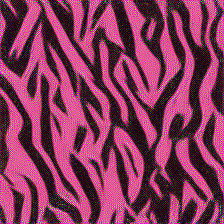} &
\includegraphics[width=\textureimgs, height=\textureimgs]{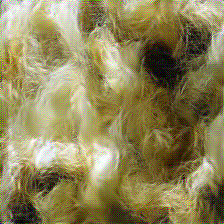} &
\includegraphics[width=\textureimgs, height=\textureimgs]{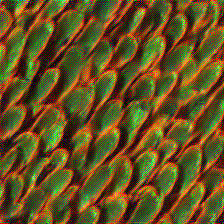} &
\includegraphics[width=\textureimgs, height=\textureimgs]{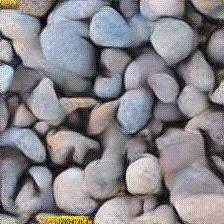}
\\ \vspace{-0.18cm}
ResNet50 &
\includegraphics[width=\textureimgs, height=\textureimgs]{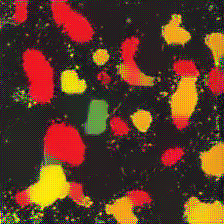} &
\includegraphics[width=\textureimgs, height=\textureimgs]{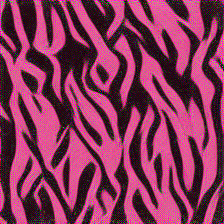} &
\includegraphics[width=\textureimgs, height=\textureimgs]{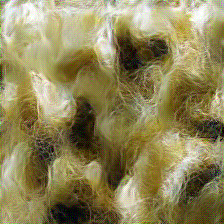} &
\includegraphics[width=\textureimgs, height=\textureimgs]{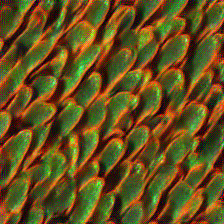} &
\includegraphics[width=\textureimgs, height=\textureimgs]{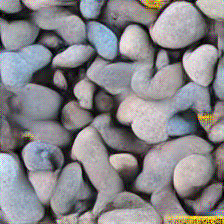}
\\\vspace{-0.18cm} 
ResNet101 &
\includegraphics[width=\textureimgs, height=\textureimgs]{figures/resnet101_blotchy_reco_s5998.png} &
\includegraphics[width=\textureimgs, height=\textureimgs]{figures/resnet101_striped_reco_s5998.png} &
\includegraphics[width=\textureimgs, height=\textureimgs]{figures/resnet101_matted_reco_s5998.png} &
\includegraphics[width=\textureimgs, height=\textureimgs]{figures/resnet101_scaly_reco_s5998.png} &
\includegraphics[width=\textureimgs, height=\textureimgs]{figures/resnet101_pebble_reco_s5998.png}
\\ \vspace{-0.18cm}
ResNet152 &
\includegraphics[width=\textureimgs, height=\textureimgs]{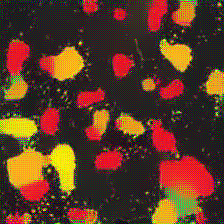} &
\includegraphics[width=\textureimgs, height=\textureimgs]{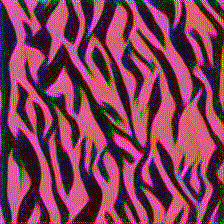} &
\includegraphics[width=\textureimgs, height=\textureimgs]{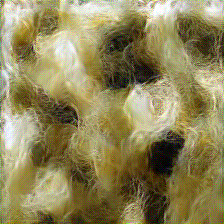} &
\includegraphics[width=\textureimgs, height=\textureimgs]{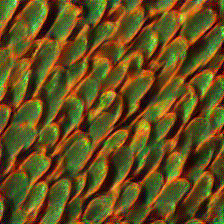} &
\includegraphics[width=\textureimgs, height=\textureimgs]{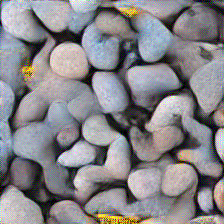}
\\ \vspace{-0.18cm}
VGG-16 &
\includegraphics[width=\textureimgs, height=\textureimgs]{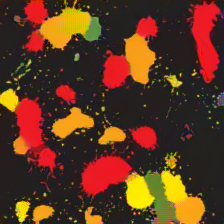} &
\includegraphics[width=\textureimgs, height=\textureimgs]{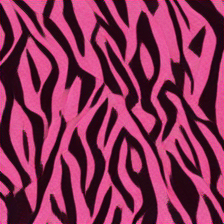} &
\includegraphics[width=\textureimgs, height=\textureimgs]{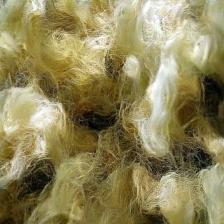} &
\includegraphics[width=\textureimgs, height=\textureimgs]{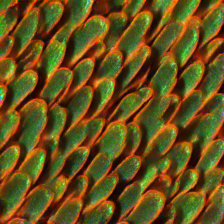} &
\includegraphics[width=\textureimgs, height=\textureimgs]{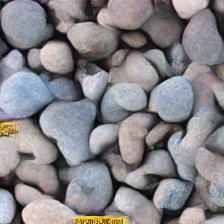}
\\ \vspace{-0.18cm}
VGG-19 &
\includegraphics[width=\textureimgs, height=\textureimgs]{figures/vgg19_blotchy_reco_s5998.png} &
\includegraphics[width=\textureimgs, height=\textureimgs]{figures/vgg19_striped_reco_s5998.png} &
\includegraphics[width=\textureimgs, height=\textureimgs]{figures/vgg19_matted_reco_s5998.png} &
\includegraphics[width=\textureimgs, height=\textureimgs]{figures/vgg19_scaly_reco_s5998.png} &
\includegraphics[width=\textureimgs, height=\textureimgs]{figures/vgg19_pebble_reco_s5998.png}
\\ 
\end{tabular}
\end{center}
\captionof{figure}{(continued from previous page)}
\label{fig:synthesized_all_2}
\end{table}
\end{document}